\documentclass[lettersize,journal]{IEEEtran}
\usepackage{amsmath,amsfonts}
\usepackage{algorithmic}
\usepackage{algorithm}
\usepackage{array}
\usepackage{textcomp}
\usepackage{stfloats}
\usepackage{url}
\usepackage{verbatim}
\usepackage{graphicx}
\usepackage{cite}
\usepackage{amssymb}
\usepackage{amsmath}
\usepackage{booktabs} 
\usepackage{algorithm}
\usepackage{algorithmic}

\usepackage{amsthm}
\PassOptionsToPackage{svgnames}{xcolor}
\usepackage[usenames,dvipsnames]{color}
\usepackage{colortbl}
\definecolor{mygray}{gray}{.9}
\usepackage{hyperref}
\usepackage{amssymb}
\usepackage{multirow} 
\usepackage{multicol}

\usepackage[ font=footnotesize ]{subfig}

\begin{document}


\title{AdaAugment: A Tuning-Free and Adaptive Approach to Enhance Data Augmentation}
\author{Suorong~Yang*, 
	Peijia~Li,
	Xin~Xiong,
	Furao~Shen*,~\IEEEmembership{Member,~IEEE}\thanks{* Corresponding author.},
        and Jian~Zhao,~\IEEEmembership{Senior Member,~IEEE}
 
	\thanks{Suorong~Yang is with State Key Laboratory for Novel Software Technology, Department of Computer Science and Technology, Nanjing University, Nanjing 210023, China (e-mail: sryang@smail.nju.edu.cn)}
	\thanks{Peijia Li, Xin Xiong, and Furao~Shen are with State Key Laboratory for Novel Software Technology, Nanjing University, China, School of Artificial Intelligence, Nanjing University, Nanjing 210023, China (e-mail: lipj@smail.nju.edu.cn; xiongxin@smail.nju.edu.cn; frshen@nju.edu.cn) }
	\thanks{Jian Zhao is with the School of Electronic Science and Engineering, Nanjing University, Nanjing 210023, China (e-mail: jianzhao@nju.edu.cn)}
}

\markboth{Journal of \LaTeX\ Class Files,~Vol.~14, No.~8, August~2021}%
{Shell \MakeLowercase{\textit{et al.}}: A Sample Article Using IEEEtran.cls for IEEE Journals}

\IEEEpubid{0000--0000/00\$00.00~\copyright~2021 IEEE}

\maketitle

\begin{abstract}
Data augmentation (DA) is widely employed to improve the generalization performance of deep models.
However, most existing DA methods employ augmentation operations with fixed or random magnitudes throughout the training process.
While this fosters data diversity, it can also inevitably introduce uncontrolled variability in augmented data, which could potentially cause misalignment with the evolving training status of the target models.
Both theoretical and empirical findings suggest that this misalignment increases the risks of both underfitting and overfitting.
To address these limitations, we propose \textbf{AdaAugment}, an innovative and tuning-free adaptive augmentation method that leverages reinforcement learning to dynamically and adaptively adjust augmentation magnitudes for individual training samples based on real-time feedback from the target network.
Specifically, AdaAugment features a dual-model architecture consisting of a policy network and a target network, which are jointly optimized to adapt augmentation magnitudes in accordance with the model's training progress effectively.
The policy network optimizes the variability within the augmented data, while the target network utilizes the adaptively augmented samples for training.
These two networks are jointly optimized and mutually reinforce each other.
 Extensive experiments across benchmark datasets and deep architectures demonstrate that AdaAugment consistently outperforms other state-of-the-art DA methods in effectiveness while maintaining remarkable efficiency.
Code is available at \url{https://github.com/Jackbrocp/AdaAugment}.
\end{abstract}


 \section{Introduction}
\IEEEPARstart{R}{ecently}, deep neural networks (DNNs) have witnessed remarkable advancements. 
Despite these advancements, achieving satisfactory performance often necessitates larger network architectures and extensive training data.
This is particularly challenging as labeled data are sometimes unavailable and expensive to collect~\cite{DA-defect-det,DA-pedestrian-det,DA-bitplane}.
Thus, the ability of deep learning models to generalize beyond the finite training set remains one of the fundamental problems in deep learning~\cite{survey3,DA-domain-generalization}. 
In this context, data augmentation (DA) emerges as a critical and highly effective technique for enhancing DNN generalization performance~\cite{survey0,survey1,survey2}, leading to the development of more generalized and robust models~\cite{investigating,DA-domain-adaptive,DA-semantic,yang2025dynamic}. 
Indeed, training large deep models to achieve state-of-the-art (SOTA) performance typically employs strong DA techniques.
\IEEEpubidadjcol

However, it is important to note that existing DA methods primarily rely on the utilization of totally random or manually devised augmentation magnitudes during training, potentially leading to suboptimal training scenarios~\cite{DA-skeletion}.
For instance, DA methods based on information deletion or fusion~\cite{cutout,gridmask,advmask,mixup,cutmix,lgcoamix,metamixup} typically erase or mix random sub-regions in images to generate augmented samples, resulting in random augmentation strengths.
Automatic DA approaches~\cite{autoaugment,randaugment,trivialaugment,selectaugment,teachaugment} often incur significant computational overhead to optimize augmentation strategies for each specific dataset prior to target model training. 
During training, the magnitudes of the applied augmentations are predetermined without any adaptation throughout the entire training process.
Moreover, most adaptive DA methods~\cite{adaptiveDA,adaaug,dada,teachaugment} select augmentation operations with random or predefined magnitudes during online training, primarily aiming to alleviate overfitting risks.
Consequently, existing DA methods may fail to capture the unique characteristics of different datasets and the entire evolving state of network training, potentially limiting their effectiveness.

\begin{figure}[]
	\centering
    \subfloat[Traditional DA]{
    \includegraphics[width=0.4\columnwidth]{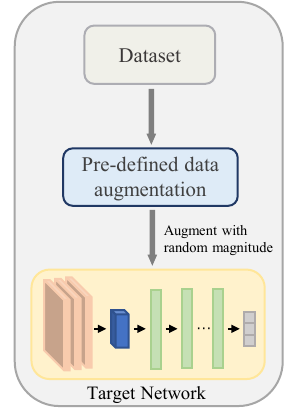}}
    \subfloat[Adaptive DA]{
    \raisebox{0.5mm}{\includegraphics[width=0.4\columnwidth]{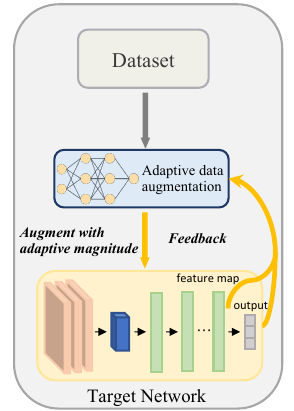}
    }}
	\caption{Comparison between traditional DA methods and our adaptive DA mechanism. (a) The typical methods employ augmentation operations with random magnitudes and may not align with the training status of deep models. In contrast, in (b), our AdaAugment adaptively determines augmentation magnitudes for each training sample based on the model's real-time feedback.}
	\label{fig-comparison-ada-da}
\end{figure}
Using random augmentation magnitudes or random operations with fixed magnitudes introduces inherent randomness, which can increase data diversity yet introduce uncontrolled variability into the augmented data. 
 Such randomness in data variability may not optimally align with the evolving training status of deep models, thereby introducing severe side effects to the training process.
 For instance, during the initial training stages, when models generally exhibit weaker generalization capabilities, substantial data variability can induce noises and distributional shifts, potentially leading to underfitting phenomena~\cite{curriculum}.
 In contrast, in the later training stages, limited data variability may elevate the risk of overfitting~\cite{advmask}.
 Thus, this misalignment in current DA methodologies increases the risks of both underfitting and overfitting, which can ultimately have adverse effects on the generalization performance of models~\cite{keepaugment,investigating}.
 \textit{This is a challenge often overlooked in the field}.

To tackle these challenges, it is crucial to adopt an adaptive modulation of augmented data variability based on real-time feedback from the target models.
As illustrated in Figure~\ref{fig-comparison-ada-da}, we present a comparison between traditional DA and adaptive DA mechanisms.
Unlike conventional approaches that rely on random or predefined augmentation magnitudes, this adaptive DA mechanism dynamically adjusts the magnitudes of DA operations based on real-time feedback from the target network.
Moreover, different from most existing adaptive DA methods, this adaptive DA mechanism \textbf{emphasizes optimizing augmentation strengths rather than the specifics of the operations themselves.}
The adaptive strategy significantly reduces the search space and offers an effective countermeasure to directly adjust the variability of the augmented data regardless of the specifics of the augmentation policies.
This brings several notable advantages: effectively mitigating the aforementioned challenges, unleashing the full potential of DA, and enhancing the generalization capabilities of deep models.
A straightforward approach to achieving adaptive DA is to create a formal measure that reflects the real-time learning status of each training sample.
This measure can then be used to determine the appropriate level of augmentation strengths, thereby adapting model training status.
Nevertheless, theoretical analyses have demonstrated that determining the model status, e.g., underfitting or overfitting risks of a learning algorithm, is undecidable~\cite{undecide,undecidability,problem-overfitting}.
Consequently, formulating a definitive measure of learning status for training samples remains a formidable challenge.

To address the aforementioned limitations and challenges, in this work, we propose \textbf{AdaAugment}, a novel and tuning-free \textbf{Ada}ptive data \textbf{Augment}ation method for image classification.
Instead of relying on any hand-crafted measures, AdaAugment leverages a reinforcement learning algorithm to adaptively determine specific augmentation magnitudes for each training sample.
Central to AdaAugment is a dual-model architecture: the policy network and the target network.
The policy network learns the policy that determines the magnitudes of augmentation operations for each training sample based on the real-time feedback from the target network during training. 
The target network simultaneously utilizes these adaptively augmented samples for training.
Both networks are jointly optimized, eliminating the need for separate re-training of the target network and enhancing the practical applicability of our approach across diverse datasets.
The learned policy adaptively adjusts the variability within augmented data to align with the learning status of the target model, thereby optimizing the introduced variability into the augmented data.
Specifically, in the training of the policy network, we estimate the risks of underfitting and overfitting by deriving losses from fully augmented and non-augmented data, respectively. 
These two losses are then compared with those derived from data adaptively augmented by AdaAugment, serving as reward signals.
In this way, \textbf{AdaAugment is capable of adjusting the emphasis to align with model training status, effectively achieving model- and data-adaptive augmentation without requiring any prior knowledge about specific datasets}.
Extensive experimental results across various benchmark datasets, including coarse-grained classification datasets CIFAR-10/100~\cite{cifar100}, Tiny-ImageNet~\cite{tiny}, and ImageNet-1k~\cite{imagenet}, long-tail classification datasets ImageNet-LT and Places-LT~\cite{openlongtailrecognition}, and several fine-grained datasets~\cite{stanford-cars,oxford-flower,oxford-pets,oxford-aircraft} demonstrate the superior effectiveness of AdaAugment in enhancing model generalization performance. 
Furthermore, complexity analyses of AdaAugment verify that it merely imposes minimal parameters and computational overhead, highlighting its competitive efficiency.
Thus, AdaAugment achieves a commendable balance between effectiveness and efficiency, yielding high performance without introducing excessive computational complexity.
For instance, AdaAugment can achieve more than 1\% improvements over other state-of-the-art DA methods when training on Tiny-ImageNet without any architectural modifications or additional regularization.

In summary, we highlight our contributions as follows:
\begin{enumerate}
    \item We propose AdaAugment, an innovative and tuning-free adaptive DA approach that leverages reinforcement learning techniques to dynamically adapt augmentation magnitudes for individual training samples based on real-time feedback from the target network.
    \item AdaAugment features a dual model architecture that jointly optimizes a policy network and a target network. This facilitates policy formulation and enhances task performance concurrently and effectively.
    \item AdaAugment employs both model- and data-adaptive DA operations to dynamically adapt model training progress.
    \item Extensive experiments conducted on multiple benchmark datasets and deep architectures demonstrate that AdaAugment outperforms existing SOTA DA methods with competitive training efficiency. This creates a strong baseline of data augmentation for future research.
\end{enumerate}

 \section{Related Work}
\subsection{Data Augmentation}
DA has played a crucial role in enhancing the generalization capabilities of deep neural networks.
Existing DA methods could be broadly divided into two categories: methods based on information deletion or fusion~\cite{gridmask,guidedmixup,lgcoamix,mixup,cutmix,cutout,advmask,random_erasing} and automatic DA methods~\cite{autoaugment,randaugment,trivialaugment,selectaugment,adaptiveDA,teachaugment,adaaug,dada,entaugment}.
Typically, these methods have primarily relied on augmentations characterized by random or predefined magnitudes to introduce diversity into the training data.

\textbf{Methods based on information deletion or fusion.} 
Among these methods, Cutout~\cite{cutout} is one of the most widely used techniques that randomly masks out one or more square regions within images.
Random Erasing~\cite{random_erasing} randomly selects a rectangular region within an image and erases its pixels with random values.
Similarly, Hide-and-Seek (HaS)~\cite{has} randomly hides patches in a training image, improving object localization accuracy and the generalization ability of deep models.
GridMask~\cite{gridmask} employs structured dropping regions within the input images.
Since these methods may easily introduce noise and ambiguity into the augmented data, AdvMask~\cite{advmask} identifies classification-discriminative information in images and structurally drops some sub-regions containing critical points for augmentation.
Additionally, Mixup~\cite{mixup} and CutMix~\cite{cutmix} synthesize augmented data by mixing random information from two or more images.
However, these DA approaches primarily focus on data transformations and often overlook the training status of models.
This oversight makes it challenging to alleviate overfitting risks accordingly through online adjustments to augmentation strengths.

\textbf{Automatic DA methods.} This group of studies can be broadly divided into fixed augmentation policies and online augmentation policies.
Methods based on fixed augmentation policies, such as AutoAugment~\cite{autoaugment}, Fast-AutoAugment~\cite{fast_autoaugment}, Faster-AutoAugment~\cite{faster}, and RandAugment~\cite{randaugment} leverage reinforcement learning or grid search to search existing policies for the optimal combination of DA operations on different image datasets in an offline manner. 
Similarly, Adversarial AutoAugment~\cite{adversarial-autoaugment} utilizes a fixed augmentation space and rewards the policies that yield the lowest accuracy, causing the policy distribution to shift towards progressively stronger augmentations throughout training.
TrivialAugment~\cite{trivialaugment} employs the same augmentation space obtained by these automatic DA methods and applies a single augmentation operation to each image during training.
MetaAugment~\cite{metaaugment} leverages a static augmentation space, and its augmentation policy network outputs weights for augmented data loss.
However, the magnitudes of augmentation policies employed in these methods remain fixed or are randomly sampled during online training, leading to the uncontrollable extent of data transformations~\cite{madaug}.

On the other hand, methods based on online policies focus on adaptively selecting augmentation policies during training through various optimization processes~\cite{adaptiveDA}.
Meanwhile, they primarily aim to alleviate overfitting risks. 
For instance, the work~\cite{adaptiveDA} proposes an adaptive DA method that focuses on identifying small transformations that yield maximal classification loss on the augmented data.
SelectAugment~\cite{selectaugment} employs hierarchical reinforcement learning to acquire online policies that determine the portion of the augmented data and whether each individual sample should be augmented. 
Since the specific augmentation transformation applied to each sample is AutoAugment, Mixup, or CutMix, the DA strengths employed are still uncontrollable.
The work~\cite{learningDA} leverages impact modeling via the influence function to learn differentiable DA transformation.
The work~\cite{UADA} randomly determines the types and magnitudes of DA operations for batch-level data and updates DA's parameters along the gradient direction of the target network loss.
The work~\cite{adda} proposes AdDA for contrastive learning, which allows the network to adjust the augmentation compositions and achieve more generalizable representations.
KeepAugment~\cite{keepaugment} detects and preserves the salient regions of the images during augmentation.
MADAug~\cite{madaug} trains a model-adaptive policy through a bi-level optimization scheme, minimizing a validation set loss of a model trained using the policy-produced data augmentations.
Adaaug~\cite{adaaug} learns class-dependent and potentially instance-dependent augmentations via a differentiable workflow.
Meanwhile, MADAug and Adaaug search not only the applied magnitudes but also the probability of augmentation and specific augmentation policies.
TeachAugment~\cite{teachaugment} transforms data so that they are adversarial to the target model.
More recently, the work~\cite{investigating} evaluates DA's efficacy using similarity and diversity measures, revealing their varying importance across datasets and suggesting the merit of adaptively adjusting augmentation strengths for an optimal balance.
Different from existing DA approaches, our proposed AdaAugment merely optimizes the real-value augmentation magnitudes regardless of specific augmentation policies.
This brings \textbf{several notable advantages}: the introduced data diversity can be directly optimized regardless of specific augmentation operations; optimizing real-value magnitudes significantly enhances optimization efficiency; and AdaAugment is capable of adapting DA operations with model training status.


\subsection{Reinforcement Learning}
Reinforcement learning (RL) entails learning a sequence of actions within an interactive environment by trial and error that maximizes the expected reward~\cite{rl,rl2}.
RL obtains a wide range of applications, such as autonomous driving~\cite{rl-auto-driving} and recommender systems~\cite{rl-recommender}, etc.
In the realm of RL algorithms, there are two fundamental categories: value optimization and policy optimization methods.
Value optimization methods primarily focus on the estimation of an optimal value function, which subsequently serves as a basis for deriving the optimal policy~\cite{value-based}.
Conversely, policy optimization methods estimate the optimal behavior policy without directly estimating the value functions~\cite{rl-policy-optimization}.
Moreover, the actor-critic framework, widely adopted in RL, combines the strengths of both value-based and policy-based RL methods~\cite{rl-ac,sac,ddpg}.
This framework comprises two components: the actor, responsible for learning the policy function, and the critic, tasked with evaluating the actor's chosen actions by estimating the value function.
This dual mechanism ensures more stable and efficient learning~\cite{rl-a2c2}, such as Advantage Actor-Critic (A2C)~\cite{rl-a2c}.

\section{AdaAugment}
\begin{figure*}[]
	\centering
         \includegraphics[width=0.8\textwidth]{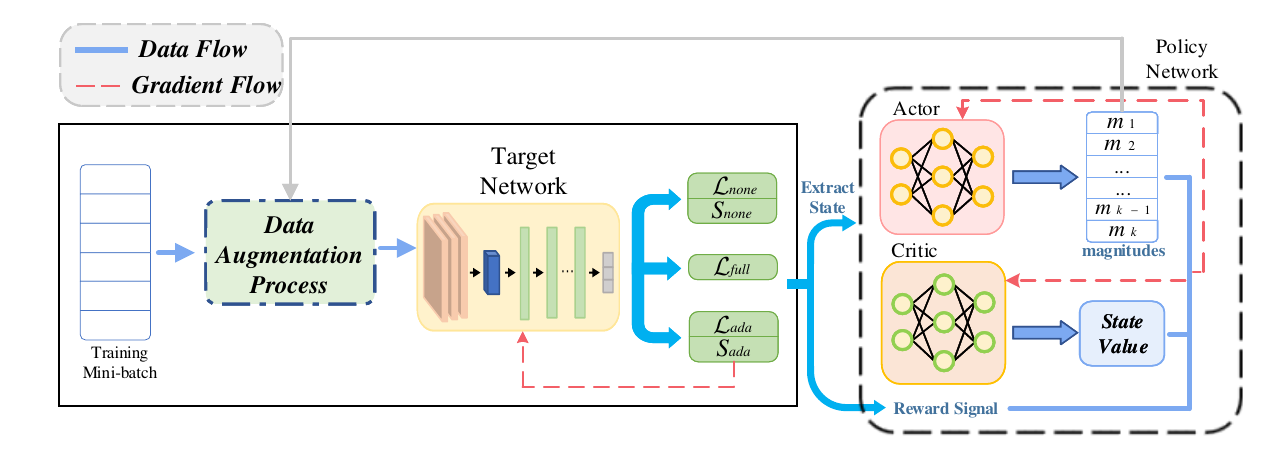}
	\caption{The dual-model framework of the proposed AdaAugment.}
    \label{fig:intro}
\end{figure*}
\paragraph{Overview}  
The primary objective of AdaAugment is to mitigate both underfitting and overfitting risks through adaptively adjusting DA strengths during the training phase. 
Specifically, it accelerates model convergence in early training stages and alleviates overfitting risks in later training stages.
This adaptive adjustment can be formalized as a decision-making problem on a per-sample basis, thereby eliminating the need for manually crafted metrics.
Figure~\ref{fig:intro} illustrates the dual-model framework of AdaAugment:
While training the target network, a policy network is simultaneously introduced to dynamically optimize the magnitudes of DA operations within the augmentation process.
This dual-model framework facilitates the joint optimization of both networks, eliminating the need for separate re-training of the target network and enabling real-time adjustments of augmentation strengths based on the training progress of the target network.
Specifically, the policy network optimizes magnitudes at the sample level in the current training epoch.
Subsequently, in the next epoch, the data augmentation process applies augmentation operations to the training samples with these corresponding magnitudes.
The adaptively augmented samples are then employed to optimize the target network parameters. 

\paragraph{Preliminary} An RL task can be formalized as a Markov Decision Process (MDP), consisting of the following components: a state space $\mathcal{S}$, an action space $\mathcal{A}$, a transition function $\mathcal{P}:\mathcal{S} \times \mathcal{A} \times \mathcal{S} \rightarrow[0,1]$ representing the probability of transitioning from one state to another by taking actions, a reward function $\mathcal{R}$, a discount factor $\gamma \in [0,1]$, and a time step $T$.
 Given a state $\boldsymbol{s} \in \mathcal{S}$, an RL agent determines an action $a\in \mathcal{A}$ with the policy $\pi(a \mid \boldsymbol{s})$.
 With these elements, the objective of an RL task is to find an optimal policy, denoted as $\pi^*$, that maximizes the expected cumulative reward within the given MDP framework.
 \begin{table}[]
 \centering
        \caption{The Augmentation Space $\mathcal{E}$.}
 \renewcommand\arraystretch{1.}
 \resizebox{.65\columnwidth}{!}
 {
  \begin{tabular}{c|c|c}
   \toprule[1.2pt]
           Transformation & $S_{Max}$ & Symmetric  \\
           \hline
           identity & - & - \\
           auto contrast & - & - \\
           equalize & - & - \\
           color & 1.9 & - \\
           contrast & 1.9 & - \\
           brightness & 1.9 & - \\
           sharpness & 1.9 & - \\
           rotation & $30^\circ$ & $\pm$ \\
           $\text{translate}_x$ &10 & $\pm$ \\
           $\text{translate}_y$ &10 & $\pm$ \\
          $\text{shear}_x$ & 0.3 & $\pm$ \\
           $\text{shear}_y$ & 0.3 & $\pm$  \\
            solarize & 256 &- \\
            posterize & 4 & - \\
           
   \bottomrule[1.2pt]
  \end{tabular}
  }
 \label{tab:augmentation-space}
\end{table}

Suppose that the training dataset $\mathcal{D}$ consists of $N$ training samples, each of the form $(\boldsymbol{x}, \boldsymbol{y}) \in \mathcal{D}$. 
 Here, $\boldsymbol{x}$ denotes the original data, and $\boldsymbol{y}$ is a $k$-dimensional vector of zeros and ones indicating the true label of $\boldsymbol{x}$, where $k$ is the total number of classes.
 The augmentation operation is defined as $e(m,\boldsymbol{x})$, where $e$ is randomly drawn from an augmentation space $\mathcal{E}$ and $m$ represents the corresponding magnitude of $e$.
Although $e$ is randomly selected from $\mathcal{E}$, the applied magnitudes are adaptively determined by AdaAugment.
This effectively adjusts the augmented data variability,  achieving a superior balance between the diversity and consistency of the training data, leading to improved model generalization.
 
 We provide the specific details of $\mathcal{E}$ in Table~\ref{tab:augmentation-space}, which contains widely used image operations and their corresponding maximal magnitudes.
 Notably, the augmentation space $\mathcal{E}$ closely follows the setting from~\cite{trivialaugment,autoaugment,randaugment,fast_autoaugment}.
 Unlike previous works, our approach adaptively determines the magnitude in the augmentation space $\mathcal{E}$ rather than assigning a predefined value before training.
 In this way, the applied magnitudes are determined by $S_{Max}\times m$, where $S_{Max}$ is the maximum allowable strength of the corresponding transformations.
 Moreover, for symmetric transformations, the symmetric direction is randomly selected.

\paragraph{State Design} 
 Since the objective of RL is to dynamically determine the appropriate magnitudes for each sample, the state $\boldsymbol{s} \in \mathcal{S}$ should consider three factors: the inherent difficulty associated with each sample, the current training status of the models (e.g., feature extraction capabilities), and the intensity of augmentation operations w.r.t. the former two factors.
 This multifaceted consideration of state variables is essential for effective RL.
 Notably, the feature map plays a vital role in this process by providing feedback from the model~\cite{feature_map,feature_map2}, encapsulating both the inherent difficulty of the sample and the model's real-time feature extraction capability.
 To illustrate this, as shown in Figure~\ref{fig:intro}, the state vector $\boldsymbol{s}$ in AdaAugment encodes the feature maps of both the non-augmented sample $\boldsymbol{x}$ and adaptively augmented sample $\Tilde{\boldsymbol{x}}$, denoted as $\boldsymbol{s}_{none}$ and $\boldsymbol{s}_{ada}$, respectively.
 The feature maps are obtained from the output of the final linear layer before the classification head in the target network.
 Here, $\boldsymbol{s}_{none}$ captures the inherent difficulty of sample $\boldsymbol{x}$ w.r.t. the current target network training status.
 Meanwhile, $\boldsymbol{s}_{ada}$ encodes the augmentation operation strength applied to $\Tilde{\boldsymbol{x}}$ w.r.t. both the real-time target network status and the inherent sample difficulty.
 By utilizing this information, AdaAugment effectively aligns data variability with the evolving training status of the target networks.

 \begin{table}[]
\centering
\caption{Network Architecture of A2C.\label{tab:a2c-arch}}
\begin{tabular}{c|c|c|c} \hline
\toprule[1.5pt]
& index & Layer & Dimension\\ \hline
\multirow{3}{*}{Actor} & 1 & Linear & (512, 512) \\ \cline{2-4}
~ & 2 & Linear & (512, 256) \\ \cline{2-4}
~ & 3 & Linear & (256, 1)\\ \hline
\multirow{3}{*}{Critic} & 1 & Linear & (512, 512) \\ \cline{2-4}
~ & 2 & Linear & (512, 256) \\ \cline{2-4}
~ & 3 & Linear & (256, 1)\\ 
\bottomrule[1.5pt]
\end{tabular}
\label{tab1}
\end{table}

\paragraph{Action Design} 
The policy determines the augmentation magnitudes $\boldsymbol{m}$ for the augmented data.
Despite the randomness in the composition of each mini-batch during training, the magnitudes $\boldsymbol{m}$ operate on a per-sample basis, corresponding to each training sample.
For simplicity, we denote the magnitudes of the current mini-batch data as $\boldsymbol{m}$, where the dimension of $\boldsymbol{m}$ equals the batch size, and each $m\in\boldsymbol{m}$ is strictly bounded within the interval $[0,1]$.
When $m=0$, no augmentation is applied, while $m=1$ indicates the maximum magnitudes for the corresponding augmentation operations.
Thus, magnitudes closer to 0 yield samples that are more similar to the original data, while magnitudes closer to 1 produce more diverse data.
The adaptively augmented sample $\Tilde{\boldsymbol{x}}$ can be obtained by:
\begin{equation}\label{eq:ada_sample}
    \Tilde{\boldsymbol{x}} = e(m,\boldsymbol{x}),
\end{equation}
where $e$ is a random augmentation operation and $m$ is determined based on the action policy.
Eq.~\eqref{eq:ada_sample} enables us to optimize the similarity-diversity preference of the augmented data.
Thus, the policy network does not know the augmentation types and directions but just optimizes the magnitudes. 
More importantly, this adjustment on the similarity-diversity preference reflects AdaAugment's effectiveness in accelerating model convergence in the early training stages and alleviating overfitting risks at later training stages.

\paragraph{Reward Function}
Consider a target classification model $f_{\theta}$, parameterized by $\theta$, an input sample $\boldsymbol{x} \in \mathbb{R}^n$, and $f_{\theta}(\boldsymbol{x})$ denotes the network output.
Let $\mathcal{L}$ denote the cross-entropy loss, and $\mathcal{L}(f_{\theta}(\boldsymbol{x}),\boldsymbol{y})$ represent the loss item for the original sample $(\boldsymbol{x},\boldsymbol{y})$.
Our proposed method aims to promote model fitting in the early training stages and alleviate overfitting at later training stages by controlling the magnitudes of data augmentation operations based on feedback from the target model.
To this end, we define three loss terms based on our augmentation strategy. 
Firstly, $\mathcal{L}_{full}(f_{\theta}(\boldsymbol{x}^+), \boldsymbol{y})$ represents the loss for the sample $\boldsymbol{x}$ with the maximal augmentation magnitude, i.e., $\boldsymbol{x}^+=e(m=1, \boldsymbol{x})$.
Secondly, $\mathcal{L}_{none}(f_{\theta}(\boldsymbol{x}^-), \boldsymbol{y})$ denotes the loss for the non-augmented sample, i.e., $\boldsymbol{x}^-=e(m=0, \boldsymbol{x})=\boldsymbol{x}$.
Lastly, based on Eq.~\eqref{eq:ada_sample}, the loss for the adaptively augmented data is denoted as $\mathcal{L}_{ada}(f_{\theta}(\Tilde{\boldsymbol{x}}, \boldsymbol{y}))$, with the augmentation magnitude determined by the actor network.
Inspired by the curriculum learning~\cite{curriculum}, we formulate the reward function as follows:
\begin{equation}\label{eq:reward}
    r=\lambda (\mathcal{L}_{full}-\mathcal{L}_{ada}) + (1-\lambda)(\mathcal{L}_{ada}-\mathcal{L}_{none}),
\end{equation}
 where $\lambda \in [0,1]$ is an adjusting factor, which is initialized at 1 and gradually decreases to 0 during the training process.
Through this adaptive mechanism, we encourage the RL module to apply weaker DA at the beginning of training to accelerate model learning and progressively increase the DA magnitude in later stages to enhance sample diversity and mitigate the risk of overfitting.

 \begin{algorithm}[t]
\caption{The general workflow of AdaAugment.}\label{alg:adaaugment}
\begin{algorithmic}[1]
\REQUIRE {dataset $\mathcal{D}$, batch size $B$, an augmentation space $\mathcal{E}$, cross-entropy loss $\mathcal{L}$, total number of epochs $T$.}
\STATE Randomly initialize $f_\theta$, $\theta_a$, and $\theta_c$
\STATE $\boldsymbol{m} \leftarrow \boldsymbol{0}$
\FOR{$t=0$: $T-1$}
    \STATE Sample a mini-batch $\{\boldsymbol{x}_i, y_i\}_{i=1}^B$ from $\mathcal{D}$
    \STATE Sample a random augmentation operation $e$ from $\mathcal{E}$
    \STATE Generate the adaptively augmented sample $\Tilde{\boldsymbol{x}}_i$ with corresponding magnitude $m \in \boldsymbol{m}$ and fully augmented sample $\boldsymbol{x}_i^+$ according to Eq.~\eqref{eq:ada_sample}
    \STATE Compute the loss items $\mathcal{L}_{none}(f_{\theta}(\boldsymbol{x}_i), \boldsymbol{y}_i)$, $\mathcal{L}_{full}(f_{\theta}(\boldsymbol{x}^+_i), \boldsymbol{y}_i)$, and $\mathcal{L}_{ada}(f_{\theta}(\Tilde{\boldsymbol{x}}_i, \boldsymbol{y}_i))$
    \STATE Determine the corresponding adaptive magnitudes $\boldsymbol{m}$ using $\theta_a$
    \STATE Calculate reward value according to Eq.~\eqref{eq:reward}
    \STATE Update $\theta_a$ and $\theta_c$ according to Eq.~\eqref{loss1} and Eq.~\eqref{loss2}
    \STATE Update $f_\theta$ based on $\mathcal{L}_{ada}(f_{\theta}(\Tilde{\boldsymbol{x}}_i, \boldsymbol{y}_i))$
\ENDFOR
\ENSURE $f_\theta$
\end{algorithmic}
\end{algorithm}
\paragraph{Policy Learning} 
The policy aims to determine the instance-level magnitudes for augmentation operations.
For policy learning, we employ the widely-used A2C algorithm~\cite{rl-a2c} due to its lightweight architecture.
The details of the A2C architecture are presented in Table~\ref{tab:a2c-arch}, which consists of an actor network $\theta_a$ and a critic network $\theta_c$, which only contains several linear layers.
The actor network learns the policy, i.e., the probability distribution over actions given a particular state, $\pi(a \mid \boldsymbol{s})$.
Meanwhile, the critic network estimates the value associated with a particular state, denoted as $V_{\theta_c}(s)$.

To update the actor and critic networks, we reformulate the loss functions tailored to our specific problem scenario.  
Specifically, the loss function for updating $\theta_a$ is defined as:
\begin{equation}\label{loss1}
    \mathcal{L}_{\text{actor}} = - \log \pi_{\theta_a}(a \mid s_{none})(r+\gamma  V_{\theta_c}(s_{ada}) - V_{\theta_c}(s_{none})).
\end{equation}
 Meanwhile, the loss function for updating $\theta_c$ is defined as:
 \begin{equation}\label{loss2}
     \mathcal{L}_{\text{critic}} = \mathbb{E}[(r + \gamma V_{\theta_c}(s_{ada}) - V_{\theta_c}(s_{none}))^2].
 \end{equation}
 To provide a comprehensive insight into AdaAugment, Algorithm~\ref{alg:adaaugment} offers a detailed algorithmic procedure.

 \begin{figure}[]
	\centering
         \includegraphics[width=6cm]{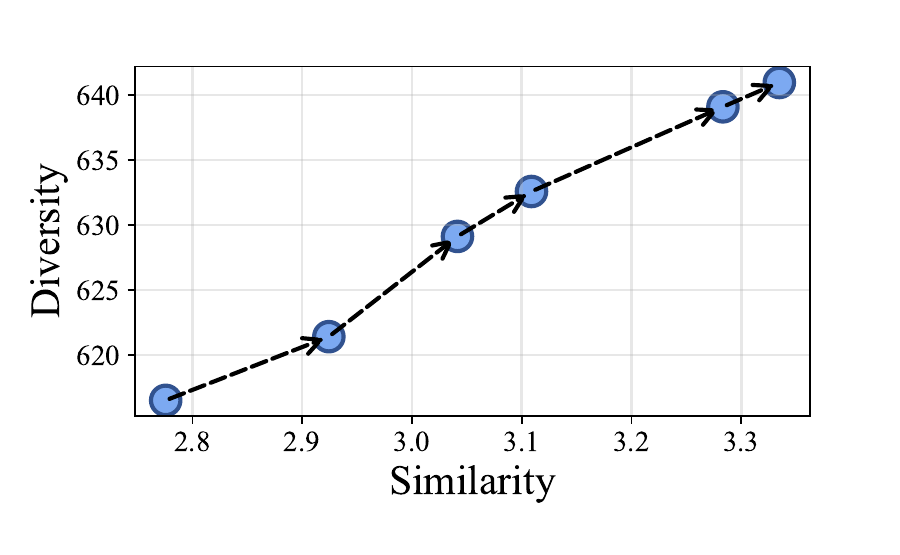}
	\caption{Evaluation of the adaptively augmented datasets throughout the training process with CIFAR-10 using ResNet-50.}
    \label{fig:mag}
\end{figure}
\paragraph{Theoretical Analysis}
In the realm of data augmentation, the augmentation magnitude is approximately proportional to the loss values of the augmented samples, i.e., $m \propto \mathcal{L}$.
Therefore, the following inequality approximately holds: $\mathcal{L}_{none} \leq \mathcal{L}_{ada} \leq \mathcal{L}_{full}$, which suggests that $\mathcal{L}_{none}$ and $\mathcal{L}_{full}$ can serve as real-time indicators of potential overfitting and underfitting status associated with augmented samples, respectively.
 Meanwhile, these loss values continuously evolve as training progresses.

Regarding the reward function in Eq.~\eqref{eq:reward}, the term $\mathcal{L}_{full}-\mathcal{L}_{ada}$ holds significant importance during the initial training phases.
Consequently, the policy network tends to employ relatively modest augmentation magnitudes, aiming to maximize the disparity between $\mathcal{L}_{full}$ and $\mathcal{L}_{ada}$.
Therefore, more similar augmented samples are encouraged initially, enabling models to capture broader features or patterns and facilitating rapid convergence during early training stages~\cite{liu2022data, madaug}. 
 As training advances, the focus shifts towards $\mathcal{L}_{ada}-\mathcal{L}_{none}$, prompting the policy network to apply greater augmentation magnitudes, trying to increase the difference between $\mathcal{L}_{ada}$ and $\mathcal{L}_{none}$.
This yields more diverse augmented samples, thereby mitigating overfitting risks.
Thus, the methodology aligns with the curriculum learning principle~\cite{curriculum,curriculum2}, wherein models initially learn patterns from similar samples and subsequently benefit from diverse samples in later training stages.
 
To further investigate the similarity-diversity trade-off of AdaAugment throughout the training process, we employ the similarity and diversity measures~\cite{investigating} for the augmented data at various training stages.
  Specifically, we calculate the average magnitude values for all samples at the intervals of 50 epochs.
 As depicted in Figure~\ref{fig:mag}, it can be observed that the diversity measure increases and the similarity measure decreases gradually as training progresses, which aligns with our theoretical analysis.
This highlights AdaAugment's inherent adaptability in adjusting DA strengths to harmonize with the evolving training dynamics of the target models.
 \begin{table*}[t]
    \centering
	\renewcommand\arraystretch{1.}
 \caption{Test accuracy (\%) on CIFAR-10/100 (average $\pm$ std). * means results reported in previous papers. }
        \small
    \resizebox{.99\textwidth}{!}{
			\begin{tabular}{c|llll|llll}
				\bottomrule[1.5pt]
       \multirow{2}*{Method} &\multicolumn{4}{c}{\cellcolor{mygray}CIFAR-10}&\multicolumn{4}{c}{\cellcolor{mygray}CIFAR-100}\\ \cline{2-9}
         & ResNet-18  & ResNet-50  &WRN-28-10 &ShakeShake & ResNet-18  & ResNet-50  &WRN-28-10 &ShakeShake \\  \midrule
      baseline & 95.28\scriptsize{$\pm$0.14}*&95.66\scriptsize{$\pm$0.08}*&95.52\scriptsize{$\pm$0.11}*&94.90\scriptsize{$\pm$0.07}*&77.54\scriptsize{$\pm$0.19}*&77.41\scriptsize{$\pm$0.27}*&78.96\scriptsize{$\pm$0.25}*&76.65\scriptsize{$\pm$0.14}* \\
      HaS~\cite{has}&96.10\scriptsize{$\pm$0.14}*&95.60\scriptsize{$\pm$0.15}&96.94\scriptsize{$\pm$0.08}&96.89\scriptsize{$\pm$0.10}*&78.19\scriptsize{$\pm$0.23}&78.76\scriptsize{$\pm$0.24}&80.22\scriptsize{$\pm$0.16}&76.89\scriptsize{$\pm$0.33}  \\
      FAA~\cite{fast_autoaugment}&95.99\scriptsize{$\pm$0.13}&96.69\scriptsize{$\pm$0.16}&97.30\scriptsize{$\pm$0.24}&96.42\scriptsize{$\pm$0.12}&79.11\scriptsize{$\pm$0.09}&79.08\scriptsize{$\pm$0.12}&79.95\scriptsize{$\pm$0.12}&81.39\scriptsize{$\pm$0.16} \\
      DADA~\cite{dada} & 95.58\scriptsize{$\pm$0.06}&95.61\scriptsize{$\pm$0.14}&97.30\scriptsize{$\pm$0.13}*&97.30\scriptsize{$\pm$0.14}* & 78.28\scriptsize{$\pm$0.22}& 80.25\scriptsize{$\pm$0.28}&82.50\scriptsize{$\pm$0.26}*& 80.98\scriptsize{$\pm$0.15} \\
      Cutout~\cite{cutout}&96.01\scriptsize{$\pm$0.18}*&95.81\scriptsize{$\pm$0.17}&96.92\scriptsize{$\pm$0.09}&96.96\scriptsize{$\pm$0.09}* &78.04\scriptsize{$\pm$0.10}*&78.62\scriptsize{$\pm$0.25}&79.84\scriptsize{$\pm$0.14}&77.37\scriptsize{$\pm$0.28} \\
     CutMix~\cite{cutmix} &96.64\scriptsize{$\pm$0.62}*&96.81\scriptsize{$\pm$0.10}*&96.93\scriptsize{$\pm$0.10}*&96.47\scriptsize{$\pm$0.07}&79.45\scriptsize{$\pm$0.17}&81.24\scriptsize{$\pm$0.14}&82.67\scriptsize{$\pm$0.22}&79.57\scriptsize{$\pm$0.10} \\  
     MADAug~\cite{madaug}&96.49\scriptsize{$\pm$0.10}&97.12\scriptsize{$\pm$0.11}&97.48\scriptsize{$\pm$0.12}&97.37\scriptsize{$\pm$0.11}&79.39\scriptsize{$\pm$0.19}&81.40\scriptsize{$\pm$0.10}&83.01\scriptsize{$\pm$0.11}&81.67\scriptsize{$\pm$0.18}\\
     AdvMask~\cite{advmask}&96.44\scriptsize{$\pm$0.15}*&96.69\scriptsize{$\pm$0.10}*&97.02\scriptsize{$\pm$0.05}*&97.03\scriptsize{$\pm$0.12}*&78.43\scriptsize{$\pm$0.18}*&78.99\scriptsize{$\pm$0.31}*&80.70\scriptsize{$\pm$0.25}*& 79.96\scriptsize{$\pm$0.27}* \\GridMask~\cite{gridmask}&96.38\scriptsize{$\pm$0.17}&96.15\scriptsize{$\pm$0.19}&97.23\scriptsize{$\pm$0.09}&96.91\scriptsize{$\pm$0.12}&75.23\scriptsize{$\pm$0.21}&78.38\scriptsize{$\pm$0.22}&80.40\scriptsize{$\pm$0.20}& 77.28\scriptsize{$\pm$0.38} \\
   
   AutoAugment~\cite{autoaugment} &96.51\scriptsize{$\pm$0.10}*&96.59\scriptsize{$\pm$0.04}*&96.99\scriptsize{$\pm$0.06}&97.30\scriptsize{$\pm$0.11}&79.38\scriptsize{$\pm$0.20}&81.34\scriptsize{$\pm$0.29}&82.21\scriptsize{$\pm$0.17}&82.19\scriptsize{$\pm$0.19}  \\
   RandAugment~\cite{randaugment} &96.47\scriptsize{$\pm$0.32}&96.25\scriptsize{$\pm$0.06}&96.94\scriptsize{$\pm$0.13}*&97.05\scriptsize{$\pm$0.15} &78.30\scriptsize{$\pm$0.15}&80.95\scriptsize{$\pm$0.22}&82.90\scriptsize{$\pm$0.29}*&80.00\scriptsize{$\pm$0.29} \\
       
       TeachAugment~\cite{teachaugment} &96.47\scriptsize{$\pm$0.09} & 96.40\scriptsize{$\pm$0.10}&97.50\scriptsize{$\pm$0.12}& 97.29\scriptsize{$\pm$0.14}  &79.27\scriptsize{$\pm$0.17} & 80.54\scriptsize{$\pm$0.23}&82.81\scriptsize{$\pm$0.20} & 81.30\scriptsize{$\pm$0.20}\\ 
       TrivialAugment~\cite{trivialaugment} &96.28\scriptsize{$\pm$0.10}&97.07\scriptsize{$\pm$0.08}&97.18\scriptsize{$\pm$0.11}&97.30\scriptsize{$\pm$0.10}&78.67\scriptsize{$\pm$0.19}&81.34\scriptsize{$\pm$0.18}&82.75\scriptsize{$\pm$0.26}&82.14\scriptsize{$\pm$0.16}  \\ 
       RandomErasing~\cite{random_erasing} &95.69\scriptsize{$\pm$0.10} &95.82\scriptsize{$\pm$0.17}&96.92\scriptsize{$\pm$0.09}&96.46\scriptsize{$\pm$0.13}* &75.97\scriptsize{$\pm$0.11}*&77.79\scriptsize{$\pm$0.32}&80.57\scriptsize{$\pm$0.15}&77.30\scriptsize{$\pm$0.18}\\ \hline
AdaAugment&\textbf{96.75}\scriptsize{$\pm$0.06}&\textbf{97.34}\scriptsize{$\pm$0.13}&\textbf{97.66}\scriptsize{$\pm$0.07}&\textbf{97.41}\scriptsize{$\pm$0.06}&\textbf{79.84}\scriptsize{$\pm$0.27}&\textbf{81.46}\scriptsize{$\pm$0.12}&\textbf{83.23}\scriptsize{$\pm$0.23}&\textbf{82.82}\scriptsize{$\pm$0.25} \\ \bottomrule[1pt]
           \end{tabular}}
    \label{tab:cifar}
\end{table*}

\section{Experiments}
\noindent \textbf{Experimental Settings}
Following previous work~\cite{advmask,dada,trivialaugment}, we evaluate the effectiveness of our method across several benchmark datasets, including the coarse-grained classification datasets CIFAR-10/100~\cite{cifar-10}, Tiny-ImageNet~\cite{tiny}, and ImageNet-1k~\cite{imagenet}.
We then employ transfer learning to evaluate the efficacy of our method in enhancing model generalization performance and transferability. 
Moreover, we demonstrate the efficacy of AdaAugment on long-tail classification datasets, including ImageNet-LT and Places-LT~\cite{openlongtailrecognition}, as well as on several fine-grained datasets, including Oxford Flowers~\cite{oxford-flower}, Oxford-IIIT Pets~\cite{oxford-pets}, FGVC-Aircrafts~\cite{oxford-aircraft}, and Stanford Cars~\cite{stanford-cars}.
In addition, we conduct effect visualization and perform a convergence analysis of AdaAugment on CIFAR-10 throughout the training process.
We also provide a complexity analysis of AdaAugment's computational efficiency and resource utilization. 
Ablation studies are conducted to show how the components affect performance.

For baseline methods, AutoAugment and Fast-AutoAugment employ 16 operations, including Shear X/Y, Rotate, AutoContrast, Invert, Equalize, Solarize, Posterize, Contrast, Color, Brightness, Sharpness, Cutout, and Sample pairing. It randomly chooses from 25 augmentation policies on CIFAR-10/100 and 24 on ImageNet-1k. Each policy contains 2 augmentation operations. RandAugment contains 14 operations, which is the same as Table~\ref{tab:augmentation-space}, and it randomly selects $N$ operations from the augmentation space with fixed magnitudes of $M$. 
TeachAugment trains augmentation models to represent geometric augmentation and color augmentation, which can represent most operations employed in AutoAugment.
TrivialAugment employs 21 augmentation operations. Besides the ones used in Table~\ref{tab:augmentation-space}, it also uses Cutout, Invert, Flip-X/Y, Sample pairing, Blur, and Smooth.
Moreover, the probability of CutMix is 0.5 on CIFAR10/100 and 1.0 on ImageNet-1k. The probability of Cutout and AdvMask is 1.0.

\noindent \textbf{Comparison with State-of-the-arts} We compare our method with 13 most representative and commonly-used data augmentation methods, including HaS~\cite{has}, Fast-AutoAugment (FAA)~\cite{fast_autoaugment}, DADA~\cite{dada}, Cutout~\cite{cutout}, CutMix~\cite{cutmix}, MADAug~\cite{madaug}, AdvMask~\cite{advmask}, GridMask~\cite{gridmask}, AutoAugment~\cite{autoaugment}, RandAugment~\cite{randaugment}, TeachAugment~\cite{teachaugment}, TrivialAugment~\cite{trivialaugment}, and RandomErasing~\cite{random_erasing}.

\noindent \textbf{Implementation Details}
The discount factor $\gamma$ in Eq.~\eqref{loss1} and Eq.~\eqref{loss2} is set to 0.99 in our method, following~\cite{gamma-1,gamma-2,gamma-3}.
We closely follow previous works~\cite{advmask,trivialaugment} in our setup. Specifically, all images are preprocessed by dividing each pixel value by 255 and normalized by the dataset statistics. For CIFAR-10 and CIFAR-100, we utilize ResNet-18/50~\cite{resnet}, Wide-ResNet-28-10 (WRN-28-10)~\cite{wrn}, and Shake-Shake-26x32 (ShakeShake)~\cite{2017Shake}. We train 1800 epochs with cosine learning rate decay for Shake-Shake using SGD with Nesterov Momentum and a learning rate of 0.01, a batch size of 256, $1e^{-3}$ weight decay. We train 300 epochs for all other networks using SGD with Nesterov Momentum and a learning rate of 0.1, a batch size of 128, a $5e^{-4}$ weight decay, and cosine learning rate decay.
For Tiny-ImageNet, we resize the images into 64$\times$64, initialize models with ImageNet pre-trained weights, and then fine-tune models using various augmentations.
In each experiment, we perform three independent random trials.
For ImageNet-1k, following~\cite{autoaugment,trivialaugment}, ResNet-50 is used as the target network. We resize images into 224$\times$224, use the learning rate of 0.1, a batch size of 256, the Nesterov Momentum with a momentum parameter of 0.9, and a weight decay of 0.4.
Note that because of the huge computational cost, the experiment in each case is performed once.
The baseline approach involves solely padding and horizontal flipping.
For a fair comparison, all methods are implemented with the same training configurations.
If not specified, all experiments are conducted on an 8-NVIDIA-2080TI-GPU server.

\begin{table}[]
 \caption{Image classification accuracy (\%) on Tiny-ImageNet dataset (average $\pm$ std). }
		\label{tab:tiny}
	\centering
	\resizebox{0.95\columnwidth}{!}{
			\begin{tabular}{c|llll}
				\toprule[1.5pt]
				 Method & ResNet-18  & ResNet-50  & WRN-50-2 & ResNext-50  \\ \hline
     baseline &61.38\scriptsize{$\pm$0.99}&73.61\scriptsize{$\pm$0.43}&81.55\scriptsize{$\pm$1.24}& 79.76\scriptsize{$\pm$1.89}\\
     HaS~\cite{has}&63.51\scriptsize{$\pm$0.58}&75.32\scriptsize{$\pm$0.59}&81.77\scriptsize{$\pm$1.16}& 80.52\scriptsize{$\pm$1.88}\\
     FAA~\cite{fast_autoaugment}  &68.15\scriptsize{$\pm$0.70}&75.11\scriptsize{$\pm$2.70}&82.90\scriptsize{$\pm$0.92}&81.04\scriptsize{$\pm$1.92} \\
     DADA~\cite{dada}&70.03\scriptsize{$\pm$0.10}&78.61\scriptsize{$\pm$0.34}&83.03\scriptsize{$\pm$0.18} & 81.15\scriptsize{$\pm$0.34} \\
     
    Cutout~\cite{cutout}&68.67\scriptsize{$\pm$1.06}&77.45\scriptsize{$\pm$0.42}&82.27\scriptsize{$\pm$1.55}& 81.16\scriptsize{$\pm$0.78}\\
    CutMix~\cite{cutmix} &64.09\scriptsize{$\pm$0.30}&76.41\scriptsize{$\pm$0.27}&82.32\scriptsize{$\pm$0.46}&81.31\scriptsize{$\pm$1.00} \\
    MADAug~\cite{madaug} &70.16\scriptsize{$\pm$0.76}&78.62\scriptsize{$\pm$0.32}&82.38\scriptsize{$\pm$0.42} & 81.41\scriptsize{$\pm$1.26} \\
    AdvMask~\cite{advmask} &65.29\scriptsize{$\pm$0.20}&78.84\scriptsize{$\pm$0.28}&82.87\scriptsize{$\pm$0.55}&81.38\scriptsize{$\pm$1.54} \\
    GridMask~\cite{gridmask}  &62.72\scriptsize{$\pm$0.91}&77.88\scriptsize{$\pm$2.50}&82.25\scriptsize{$\pm$1.47}& 81.05\scriptsize{$\pm$1.33}\\
    AutoAugment~\cite{autoaugment}&67.28\scriptsize{$\pm$1.40}&75.29\scriptsize{$\pm$2.40}&79.99\scriptsize{$\pm$2.20}&81.28\scriptsize{$\pm$0.33} \\
    RandAugment~\cite{randaugment}  &65.67\scriptsize{$\pm$1.10}&75.87\scriptsize{$\pm$1.76}&82.25\scriptsize{$\pm$1.02}&80.36\scriptsize{$\pm$0.62} \\
    TeachAugment~\cite{teachaugment} &70.05\scriptsize{$\pm$0.57} &70.56\scriptsize{$\pm$0.44}&82.95\scriptsize{$\pm$0.13}&81.39\scriptsize{$\pm$0.97} \\
    TrivialAugment~\cite{trivialaugment} &69.97\scriptsize{$\pm$0.96}&78.41\scriptsize{$\pm$0.39}&82.16\scriptsize{$\pm$0.32} & 80.91\scriptsize{$\pm$2.26} \\
    RandomErasing~\cite{random_erasing}&64.00\scriptsize{$\pm$0.37}&75.33\scriptsize{$\pm$1.58}&81.89\scriptsize{$\pm$1.40}& 81.52\scriptsize{$\pm$1.68}\\  \hline
AdaAugment&\textbf{71.25}\scriptsize{$\pm$0.64}&\textbf{79.11}\scriptsize{$\pm$1.51}&\textbf{83.07}\scriptsize{$\pm$0.78}& \textbf{81.92}\scriptsize{$\pm$0.29}\\
    \bottomrule[1.5pt]
    \end{tabular}}
\end{table}
\subsection{Results on CIFAR-10 and CIFAR-100}
In Table~\ref{tab:cifar}, we evaluate the efficacy of AdaAugment on CIFAR-10 and CIFAR-100, employing various widely-used deep networks, such as ResNet-18/50~\cite{resnet}, Wide-ResNet-28-10~\cite{wrn}, and ShakeShake-26-32~\cite{2017Shake}.
It can be observed that AdaAugment consistently exhibits superior performance in improving the accuracy of these networks when compared to existing SOTA DA methods. 
Notably, our method demonstrates significant improvements for both datasets.
For example, AdaAugment achieves improvements on the baseline model's performance of ResNet-18, ResNet-50, Wide-ResNet-28-10, and ShakeShake-26-32 by 1.47\%, 1.66\%, 2.14\%, and 2.51\%, respectively.
While on smaller models, the performance differences among different methods are marginal, using larger models can further highlight the effectiveness of AdaAugment.
For instance, AdaAugment outperforms CutMix by nearly 1\% on CIFAR-10 and 3.3\% on CIFAR-100 with ShakeShake.
This superior performance of AdaAugment can be attributed to its adaptive adjustment of augmentation magnitudes based on real-time feedback from the target network during training.
This adjustment effectively mitigates the risks of overfitting, resulting in improved generalization capabilities.
Additionally, larger models do not necessarily lead to better performance. The baseline performance of ResNet-50 underperforms that of ResNet-18 by applying simple augmentations. This is likely due to higher overfitting risks for the larger models. However, when more advanced augmentation methods are employed, ResNet-50 consistently outperforms ResNet-18, further validating the effectiveness of DA.
\begin{table}[]
    \centering
     \caption{Image classification accuracy (\%) on Tiny-ImageNet dataset with randomly initialized ResNet-50. }
		\label{tab:tiny-r50}
    \resizebox{0.95\columnwidth}{!}{
    \begin{tabular}{ccccccc}
    \toprule[1.5pt]
         baseline& HaS & FAA&DADA&Cutout&CutMix& GridMask \\
         49.73\scriptsize{$\pm$.37}&52.05\scriptsize{$\pm$.12}&52.83\scriptsize{$\pm$.02}&53.07\scriptsize{$\pm$.28}&50.89\scriptsize{$\pm$.52}&52.31\scriptsize{$\pm$.20}&50.37\scriptsize{$\pm$.11}\\ \hline
         MADAug&AA&RA&TeachAug&TA&RE&AdaAugment\\
        54.95\scriptsize{$\pm$.24}&53.37\scriptsize{$\pm$.13}&50.63\scriptsize{$\pm$.53}&53.97\scriptsize{$\pm$.23}&53.88\scriptsize{$\pm$.03}&50.74\scriptsize{$\pm$.27}&\textbf{55.98}\scriptsize{$\pm$.12}\\
    \bottomrule[1.5pt]
    \end{tabular}}
\end{table}

\subsection{Results on Tiny-ImageNet}
In this section, we evaluate the efficacy of AdaAugment on the Tiny-ImageNet dataset using pre-trained ResNet-18/50, Wide-ResNet-50-2 (WRN-50-2), and ResNext50~\cite{resnext}, with the results summarized in Table~\ref{tab:tiny}.
AdaAugment achieves significant improvements in classification accuracy across various architectures, outperforming other widely used DA methods by a substantial margin.
Particularly, AdaAugment elevates the accuracy by 9.87\%, 5.5\%, 1.52\%, and 1.16\% on ResNet-18, ResNet-50, WRN-50-2, and ResNext-50, respectively.
Furthermore, Table~\ref{tab:tiny-r50} presents the results of training ResNet-50 on Tiny-ImageNet from scratch. Even in this scenario, AdaAugment delivers superior performance, achieving an improvement of over 1\%. 
These results highlight the effectiveness of our adaptive augmentation mechanism in enhancing both pre-trained and from-scratch training setups.
Thus, we demonstrate the superior efficacy and potential of AdaAugment for enhancing model performance.
\begin{table*}[]
     \centering
     \setlength{\tabcolsep}{2.5pt}
         \caption{Top-1 accuracy (\%) on ImageNet-1k dataset with ResNet-50. Some results are adapted from~\cite{trivialaugment,dada,teachaugment}.}\label{tab:imagenet}
	\renewcommand\arraystretch{1.}
	\resizebox{.99\linewidth}{!}{
			\begin{tabular}{cccccccccccccc}
				\toprule[1.pt]
    baseline &HaS~\cite{has} & FAA~\cite{fast_autoaugment} & DADA~\cite{dada} & Cutout~\cite{cutout} & CutMix~\cite{cutmix} & MADAug~\cite{madaug} & GridMask~\cite{gridmask} & AA~\cite{autoaugment} & RA~\cite{randaugment} & TeachAug~\cite{teachaugment} & TA~\cite{trivialaugment} &RE~\cite{random_erasing} & AdaAugment \\ \hline
  77.1 &77.2 & 77.6&77.5& 77.1  & 77.2 &\textbf{78.3} &77.9 &77.6&77.8& 78.0 &77.9 &77.3 & \textbf{78.3}\\
    \bottomrule[1.pt]
    \end{tabular}}
\end{table*}
\begin{table}[]
    \centering
     \caption{Performance (\%) on ImageNet-1k with ViT-Base, ViT-Large, and Swin-Transformer using a 4-V100-GPU server.}
		\label{tab:vit}
    \resizebox{0.65\columnwidth}{!}{
    \begin{tabular}{c|cc}
    \toprule[1.5pt]
         Model & baseline & AdaAugment \\ \hline
         ViT-B &81.46&\textbf{82.55} \\
         ViT-L &83.50 &\textbf{84.67} \\
         Swin-Transformer &85.00&\textbf{85.65} \\
    \bottomrule[1.5pt]
    \end{tabular}}
\end{table}

\begin{table*}[]
     \centering
     \setlength{\tabcolsep}{2.5pt}
         \caption{Transferred test accuracy (\%) on CIFAR-10 of various DA methods. The pretrained ResNet-50 model is trained on both CIFAR-100 (upper row) and Tiny-ImageNet (bottom row).}\label{tab:transfer-learning}
	\renewcommand\arraystretch{1.}
	\resizebox{.99\linewidth}{!}{
			\begin{tabular}{cccccccccccccc}
				\toprule[1.pt]
    baseline &HaS~\cite{has} & FAA~\cite{fast_autoaugment} & DADA~\cite{dada} & Cutout~\cite{cutout} & CutMix~\cite{cutmix} & MADAug~\cite{madaug} & GridMask~\cite{gridmask} & AA~\cite{autoaugment} & RA~\cite{randaugment} & TeachAug~\cite{teachaugment} & TA~\cite{trivialaugment} &RE~\cite{random_erasing} & AdaAugment \\ \hline
 91.53\scriptsize{$\pm$.03}&92.51\scriptsize{$\pm$.24}&92.28\scriptsize{$\pm$.13}&92.58\scriptsize{$\pm$.09}&92.42\scriptsize{$\pm$.20}&92.81\scriptsize{$\pm$.47}&92.84\scriptsize{$\pm$.10}&91.49\scriptsize{$\pm$.10}&92.82\scriptsize{$\pm$.04}&92.78\scriptsize{$\pm$.23}&92.83\scriptsize{$\pm$.18}&92.80\scriptsize{$\pm$.16}&92.55\scriptsize{$\pm$.05}&\textbf{93.06}\scriptsize{$\pm$.25}\\
 64.02\scriptsize{$\pm$.05}&66.84\scriptsize{$\pm$.06}&70.32\scriptsize{$\pm$.63}&69.04\scriptsize{$\pm$.43}&65.54\scriptsize{$\pm$.75}&69.29\scriptsize{$\pm$.09}&72.82\scriptsize{$\pm$.32}&64.88\scriptsize{$\pm$.43}&69.53\scriptsize{$\pm$.53}&64.68\scriptsize{$\pm$.97}&69.98\scriptsize{$\pm$.17}&71.53\scriptsize{$\pm$.35}&64.56\scriptsize{$\pm$.27}&\textbf{76.86}\scriptsize{$\pm$.12}\\
    \bottomrule[1.pt]
    \end{tabular}}
\end{table*}

\subsection{Results on ImageNet-1k}

 \begin{table*}[]
 \caption{Top-1 classification accuracy (\%) on ImageNet-LT and Places-LT. * means results reported in the original paper. }
		\label{tab:imagenet-LT}
    \centering
    \setlength{\tabcolsep}{2.5pt}
        \resizebox{.99\textwidth}{!}{\begin{tabular}{c|l|cccc|cccc}
          \bottomrule[1.1pt]
        \multirow{2}{*}{Dataset}& \multirow{2}{*}{Methods} &\multicolumn{4}{c}{\cellcolor{mygray}closed-set setting}&\multicolumn{4}{|c}{\cellcolor{mygray}open-set setting} \\ \cline{3-10}
        & & \textbf{Many-shot} & \textbf{Medium-shot} & \textbf{Few-shot} & \textbf{Overall}  & \textbf{Many-shot} & \textbf{Medium-shot} & \textbf{Few-shot} & \textbf{F-measure}\\ \hline
      \multirow{2}{*}{ImageNet-LT} & OLTR  & 43.2\scriptsize{$\pm$0.1}* & 35.1\scriptsize{$\pm$0.2}* & 18.5\scriptsize{$\pm$0.1}* & 35.6\scriptsize{$\pm$0.1}* &41.9\scriptsize{$\pm$0.1}*& 33.9\scriptsize{$\pm$0.1}*& 17.4\scriptsize{$\pm$0.2}* &44.6\scriptsize{$\pm$0.2}* \\ 
        &  OLTR+\textbf{AdaAugment}&\textbf{45.9}\scriptsize{$\pm$0.1} &  \textbf{38.3}\scriptsize{$\pm$0.1} & \textbf{22.0}\scriptsize{$\pm$0.2} &  \textbf{39.0}\scriptsize{$\pm$0.1} &\textbf{44.1}\scriptsize{$\pm$0.1}&\textbf{36.8}\scriptsize{$\pm$0.1}&\textbf{20.8}\scriptsize{$\pm$0.2}&\textbf{45.8}\scriptsize{$\pm$0.1}\\ 
        \hline
      \multirow{2}{*}{Places-LT } &  OLTR  & \textbf{44.7}\scriptsize{$\pm$0.1}* & 37.0\scriptsize{$\pm$0.2}* & 25.3\scriptsize{$\pm$0.1}* & 35.9\scriptsize{$\pm$0.1}*  &\textbf{44.6}\scriptsize{$\pm$0.1}*& 36.8\scriptsize{$\pm$0.1}*& 25.2\scriptsize{$\pm$0.2}* & 46.4\scriptsize{$\pm$0.1}*\\
        &  OLTR+\textbf{AdaAugment} & 43.7\scriptsize{$\pm$0.1} &  \textbf{41.1}\scriptsize{$\pm$0.1} & \textbf{29.5}\scriptsize{$\pm$0.2} &  \textbf{39.6}\scriptsize{$\pm$0.1} &43.9\scriptsize{$\pm$0.1}&\textbf{40.8}\scriptsize{$\pm$0.1}&\textbf{28.9}\scriptsize{$\pm$0.1}&\textbf{50.4}\scriptsize{$\pm$0.1}\\ 
         \bottomrule[1.1pt]
        \end{tabular}}
\end{table*}
We evaluate AdaAugment on the more challenging large-scale IamgeNet-1k dataset~\cite{imagenet}.
Note that the method AdvMask is not compared on ImageNet-1k due to its huge time costs of the sparse adversarial attack module~\cite{advmask}.
As shown in Table~\ref{tab:imagenet} and Table~\ref{tab:vit} with more advanced baseline models, on ImageNet-1k, all DA methods' average accuracy significantly outperforms the baseline.
Moreover, AdaAugment achieves the highest improvements in accuracy using ResNet-50.
Notably, AdaAugment achieves 1.1\% improvements compared to the baseline without any architecture modifications and noticeable additional training costs.
Although MADAug obtains similar results to ours on ImageNet-1k, it involves a bi-level optimization scheme that introduces much higher training costs than AdaAugment.
Moreover, in contrast to previous works, which employed fixed augmentation operations for the entire dataset, our method applies model- and sample-adaptive operations for each image.
This significantly enhances the model's generalization.


\subsection{Transfer Learning}
Transfer learning plays a pivotal role in assessing the models' transferability~\cite{transfer-learning-2}, a common practice in data augmentation~\cite{keepaugment,advmask,investigating}.
To quantitatively assess the efficacy of various DA methods, we assess the transferred accuracy of models using different DA methods.  
Specifically, we pre-train ResNet-50 models on CIFAR-100 and Tiny-ImageNet utilizing different augmentations, followed by fine-tuning these models on the CIFAR-10 dataset. 
Consequently, superior DA techniques result in higher transferred test accuracy.

In Table~\ref{tab:transfer-learning}, we present the transferred accuracy achieved by various DA methods.
While the discrepancies in transferred accuracy may appear modest, it is noteworthy that AdaAugment consistently outperforms others. 
This consistent and effective performance underscores our method's capability to enhance model transferability performance.
Thus, AdaAugment indeed enhances the model generalization performance.

\subsection{Results on Long-Tail Datasets}\label{appendix:oltr}

While most DA methods have not been tested on large-scale long-tail biased datasets, we strengthen the generality and effectiveness of our method by applying it to the ImageNet-LT and Places-LT datasets~\cite{openlongtailrecognition}.
Specifically, we utilize the codebase provided by OLTR~\cite{openlongtailrecognition} and rigorously follow all the hyperparameters and training strategies.

As shown in Table~\ref{tab:imagenet-LT}, AdaAugment brings substantial improvements across all the many/medium/few-shot classes and the open classes in both long-tail datasets.
Notably, AdaAugment achieves \textbf{over 3\% accuracy improvements in the closed-set setting}, highlighting the effectiveness of our method in improving model robustness and generalization.
Therefore, AdaAugment can also be effectively employed in more challenging long-tail datasets as well.

\subsection{Results on Fine-grained Datasets}
 \begin{table}[]
	\centering
 \caption{Test accuracy (\%) of AdaAugment on fine-grained datasets with ResNet-50. The performance is average$\pm$std.}
	\renewcommand\arraystretch{1.}
	\resizebox{0.4\textwidth}{!}{
        \begin{tabular}{l|c c}\toprule[1.5pt]
    Dataset & baseline & AdaAugment \\ \hline
    Oxford Flowers~\cite{oxford-flower} &89.47\scriptsize{$\pm$0.08}&\textbf{97.17}\scriptsize{$\pm$0.14} \\ 
    Oxford-IIIT Pets~\cite{oxford-pets}  &89.73\scriptsize{$\pm$0.18} &\textbf{91.95}\scriptsize{$\pm$0.24}  \\
    FGVC-Aircraft~\cite{oxford-aircraft}  &77.25\scriptsize{$\pm$0.09}&\textbf{90.92}\scriptsize{$\pm$0.05} \\ 
    Stanford Cars~\cite{stanford-cars}  &82.13\scriptsize{$\pm$0.03}&\textbf{84.76}\scriptsize{$\pm$0.20} \\ 
        \bottomrule[1.5pt]
    \end{tabular}}
    \label{supp:fine}
\end{table}
To comprehensively assess the effectiveness of our proposed methods, we apply AdaAugment to various fine-grained datasets, including Oxford Flowers~\cite{oxford-flower}, Oxford-IIIT Pets~\cite{oxford-pets}, FGVC-Aircraft~\cite{oxford-aircraft}, and Stanford Cars~\cite{stanford-cars}.
Specifically, for all the fine-grained datasets, we employ the ResNet-50 model~\cite{resnet} pre-trained on ImageNet, followed by fine-tuning these models using our proposed methods.
To ensure fairness, experiments on the same dataset utilize the same experimental settings.

Table~\ref{supp:fine} summarizes the test accuracy of our methods compared to the baseline. It can be observed that AdaAugment brings notable accuracy improvements across all fine-grained datasets.
Specifically, AdaAugment improves the model performance on Oxford Flowers by 7.70\%, Oxford-IIIT Pets by 2.22\%, FGVC-Aircraft by 13.67\%, and Stanford Cars by 2.63\%, respectively.
Hence, our proposed methods can also be utilized to substantially enhance the model performance on fine-grained datasets.

\subsection{Analytical Results}

\begin{figure}[]
	\centering
        \subfloat[baseline\\ DI=5.03$\times 10^{-5}$]{\label{fig2-1}
            \includegraphics[width=0.45\columnwidth]{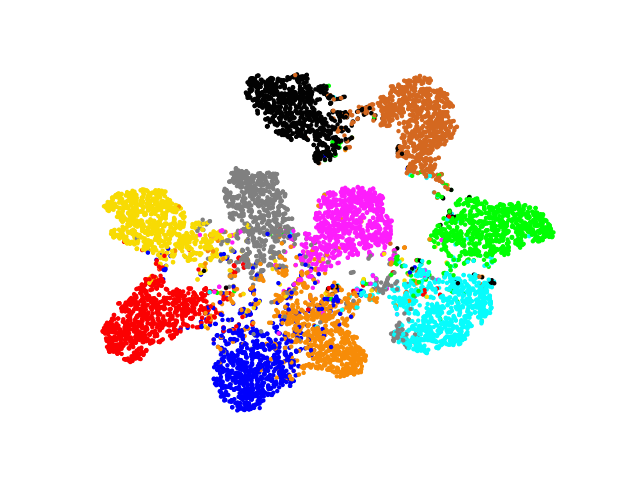}
            }  
       \subfloat[AdaAugment\\ DI=$1.1 \times 10^{-4}$]{\label{fig2-2}  
           \includegraphics[width=0.45\columnwidth]{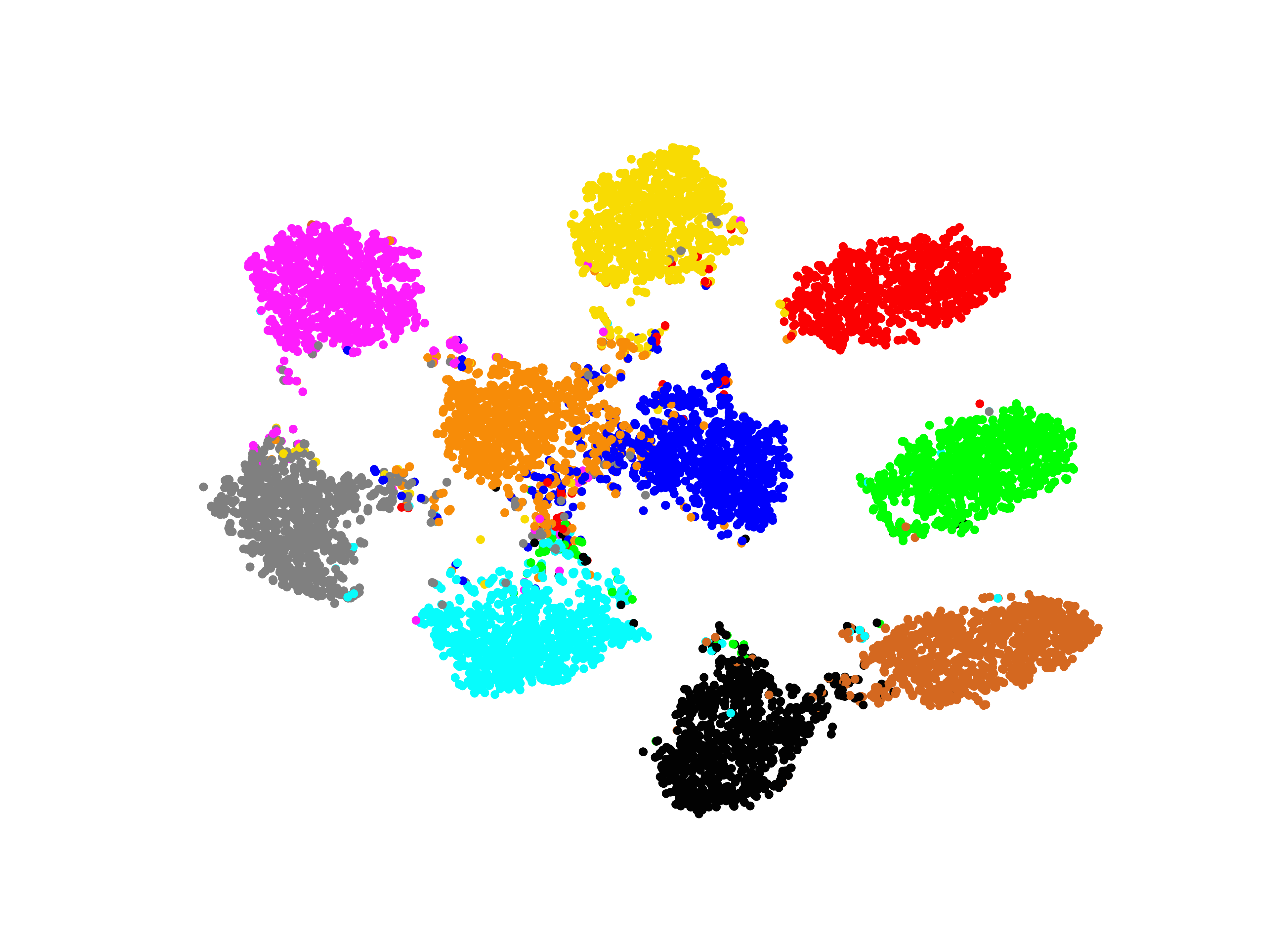}
           
               } 
	\caption{Visualization of effectiveness using t-SNE algorithm with CIFAR-10 dataset. The embedding model is ResNet-50. DI: Dunn index.}
	\label{fig-tsne}
\end{figure}
\noindent \textbf{Visualization of effectiveness.}\quad
Since the primary objective of DA is to enhance the generalization performance of models, in this section, we conduct a comparative analysis of effectiveness visualization between models utilizing AdaAugment and those without.
Specifically, we leverage the t-SNE algorithm~\cite{tsne} to visualize the feature embedding of the CIFAR-10 test set, utilizing feature maps derived from different trained models.
Theoretically, models that showcase more robust generalization capabilities tend to excel in feature extraction ability.

Figure~\ref{fig-tsne} presents the t-SNE visualization results for the baseline model and the model trained with AdaAugment.
Compared to Figure~\ref{fig-tsne}\subref{fig2-1}, it can be seen that Figure~\ref{fig-tsne}\subref{fig2-2} exhibits an optimized geometric structure characterized by \textbf{enhanced inter-cluster separation and intra-cluster compactness}.
Furthermore, to furnish a quantitative evaluation of these embedding results, we employ the Dunn Index (DI)~\cite{dunn} as an evaluative metric.
The DI is mathematically defined as follows:
\begin{equation}
    D I=\frac{\min _{1 \leq i \neq j \leq m} \delta\left(C_i, C_j\right)}{\max _{1 \leq j \leq m} \Delta_j},
\end{equation}
where separation $\delta\left(C_i, C_j\right)$ denotes the inter-cluster distance metric between clusters $C_i$ and $C_j$, and compactness $\Delta_j$ calculates the mean distance between all pairs within each cluster. 
A higher DI value signifies superior clustering performance.
The DI values for the baseline in Figure~\ref{fig-tsne}\subref{fig2-1} and AdaAugment in Figure~\ref{fig-tsne}\subref{fig2-2} are $5.03\times 10^{-5}$ and $1.05\times 10^{-4}$, respectively.
Consequently, AdaAugment achieves a DI value that is \textbf{108.7\% higher} than that of the baseline, demonstrating its superior efficacy in enhancing model performance.
Through both quantitative and qualitative analyses, we validate the effectiveness of AdaAugment in improving model performance.

\begin{table}[]
	\centering
        \caption{The additional architecture parameters and training time of the auxiliary policy network on CIFAR-10. The device used is 2 NVIDIA RTX2080TI GPUs and an Intel(R) Xeon(R) CPU E5-2678 @ 2.50GHz.}
	\renewcommand\arraystretch{1.2}
	\resizebox{.95\columnwidth}{!}{
		\begin{tabular}{c|c|ccc}
			\toprule[1.5pt]
                Models & FLOPs & Parameters &  GPU hours & Improved Acc.\\ \hline
			 ResNet-18 &1.82G & +0.15M & +0.41h\scriptsize{$\pm$0.03} & +1.47\%\scriptsize{$\pm$0.06} \\
              ResNet-50 & 4.14G & +0.60M & +0.49h\scriptsize{$\pm$0.03} & +1.68\%\scriptsize{$\pm$0.13} \\
              WRN-28-10 &5.25G & +0.19M & +0.43h\scriptsize{$\pm$0.03} & +2.04\%\scriptsize{$\pm$0.07}\\
			\bottomrule[1.5pt]
		\end{tabular}}
	\label{tab:complexity}
\end{table}

\noindent \textbf{Complexity analysis.}\quad \label{sec:complexity-analysis}
Most existing DA methods typically rely on predefined random magnitude values during training.
While this mechanism incurs minimal additional computational overhead in online training, it may not be optimally aligned with the dynamic evolution of deep models, potentially introducing undesirable side effects.
To address this, AdaAugment incorporates an auxiliary policy network into online training to dynamically determine the magnitudes of DA operations.
Although this adjustment inevitably introduces slightly additional parameters and computational complexity, it is a deliberate trade-off to achieve more effective DA.

In this section, following the methodology employed in prior studies~\cite{autoaugment,fast_autoaugment,selectaugment}, we thoroughly analyze the parameter and time complexity of AdaAugment to assess its efficiency.
Since the policy network's parameter complexity is related to the feature space of the target network, we report the parameter complexity on ResNet-18/50 and WRN-28-10 in Table~\ref{tab:complexity}.
Notably, it can be observed that \textbf{the policy network's parameter complexity results in a mere 1.3\% increase for ResNet-18} (with 11.7M parameters), \textbf{a 2.4\% increase for ResNet-50} (with 25.5M parameters), and \textbf{a 0.52\% increase for WRN-28-10} (with 36.5M parameters).
Furthermore, the overall training costs experience a slight increase for these networks.
Despite the marginal increases in training costs, AdaAugment achieves significant accuracy improvements, with a notable enhancement of 1.47\%, 1.68\%, and 2.04\% for ResNet-18, ResNet-50, and WRN-28-10, respectively.
This improvement underscores AdaAugment's remarkable effectiveness-efficiency trade-off, demonstrating its capacity to enhance model performance while incurring minimal training costs.

\begin{figure}[]
	\centering
         \includegraphics[width=8cm]{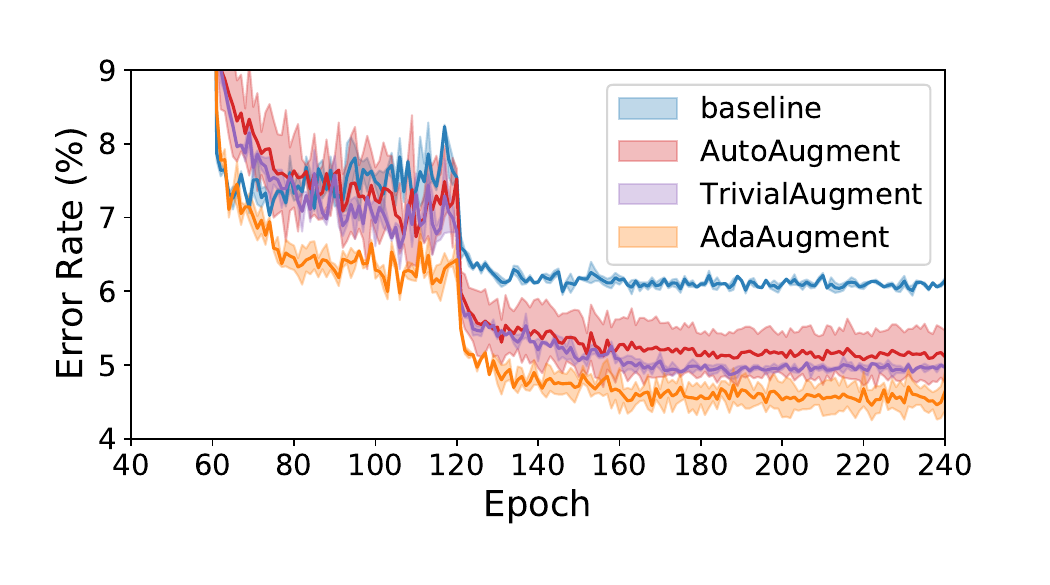}
	\caption{Convergence analysis of training process with CIFAR-10 using ResNet-50.}
    \label{fig:convergence}
\end{figure}

\begin{figure}
    \centering
    \includegraphics[width=5.5cm]{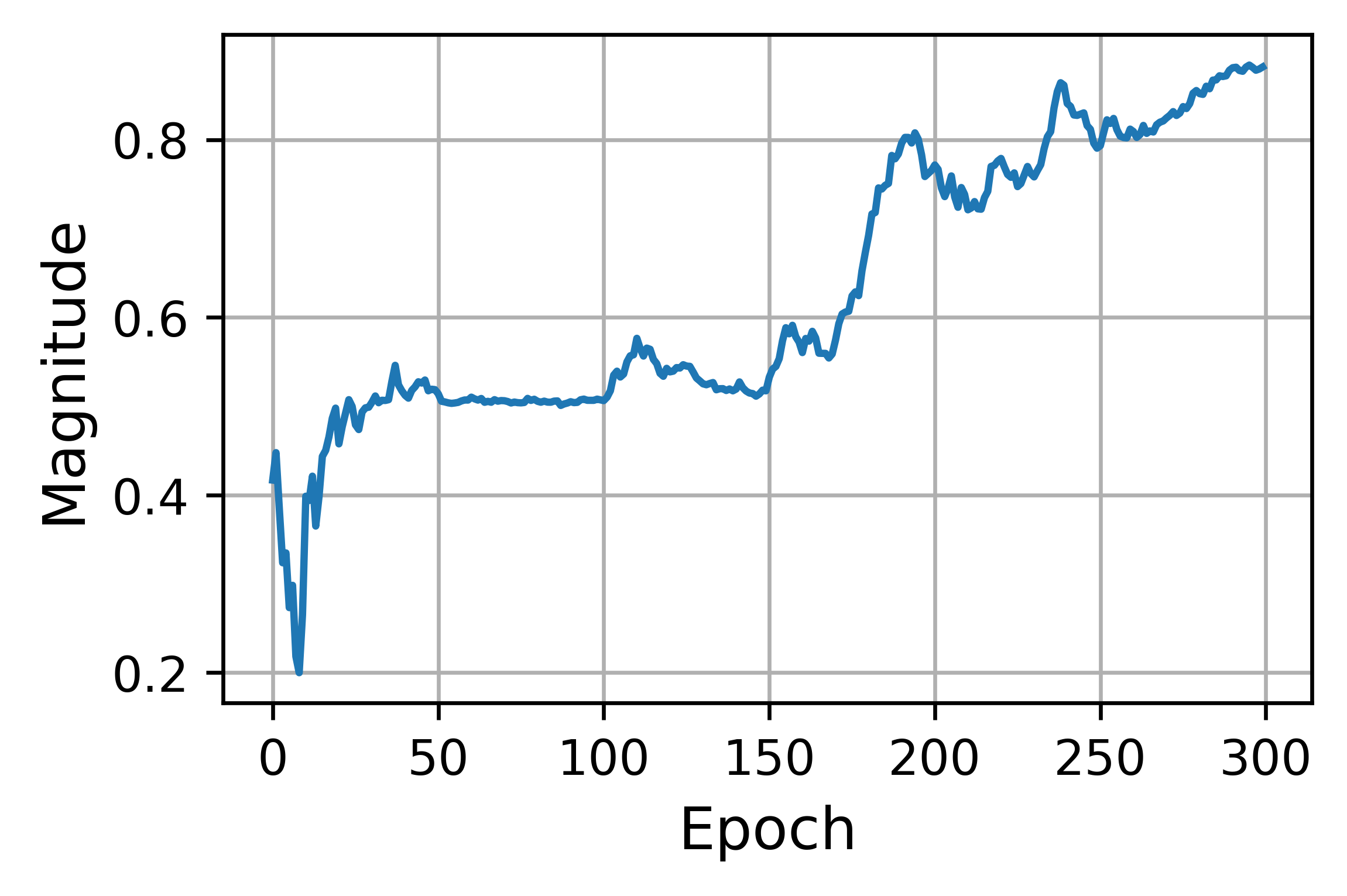}
    \caption{Evolving magnitudes throughout the training process with CIFAR-100 using ResNet-50.}
    \label{fig:magnitude-progress}
\end{figure}

\noindent \textbf{Convergence analysis.} \quad
In this section, we conduct a convergence analysis of AdaAugment throughout the entire training process, comparing it against two of the most representative DA methods, AutoAugment and TrivialAugment.
Specifically, to illustrate the dynamic evolution of test errors during training, we train ResNet-18 models on CIFAR-10 using a multi-step learning rate decay schedule.
The learning rate is initialized as 0.1 and multiplied by 0.2 at epochs 60, 120, 160, 220, and 280. 
From Figure~\ref{fig:convergence}, it can be observed that AdaAugment significantly enhances model performance, particularly after the second learning rate drop.
Moreover, even after the initial learning rate drop, AdaAugment consistently maintains lower error rates than other methods.
These empirical findings not only underscore the superior efficacy of AdaAugment in terms of model performance enhancement but also emphasize its capacity to expedite the convergence of the model during training. Therefore, AdaAugment contributes to decreasing the training difficulty of deep networks.

\noindent \textbf{Adaptively evolving magnitude.}\quad
To present the evolving progress of the adaptive magnitudes, we train ResNet-50 with CIFAR-100.
The learning rate is initialized as 0.1 and multiplied by 0.2 at epochs 60, 120, 160, 220, and 280, along with a 5-epoch scheduler warmup.
In Figure~\ref{fig:magnitude-progress}, it can be seen that the magnitude generally increases through model training, providing progressively more diverse augmented samples to mitigate the risk of overfitting.
Moreover, this dynamic alignment aligns with the model's training status. For instance, at different training stages, such as the warmup phase at the beginning of training or during learning rate changes, the magnitudes adaptively modify values, aligning with the model's evolving status.

\begin{figure}[]
	\centering
        \subfloat[Training Accuracy.]{\label{fig7-1}
        \includegraphics[width=0.45\columnwidth]{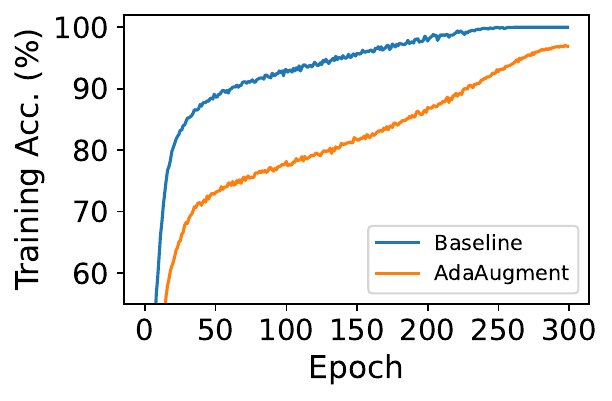}
            }  
       \subfloat[Test Accuracy.]{\label{fig7-2}  
           \includegraphics[width=0.45\columnwidth]{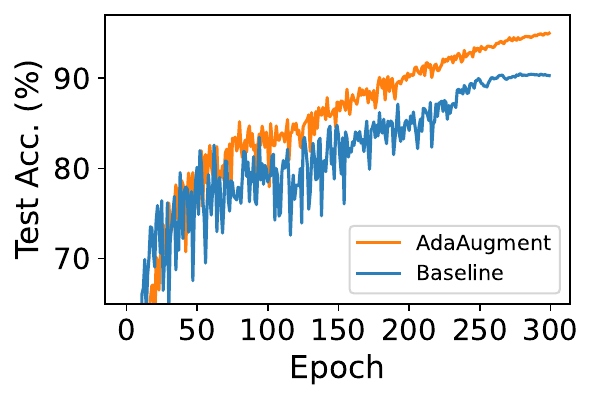}
               } 
	\caption{Effectiveness of AdaAugment in mitigating overfitting risks with reduced CIFAR-10 using ResNet-50.}
	\label{fig-overfitting-risk}
\end{figure}
\begin{table*}[]
 \centering
        \caption{Impact of the augmentation space with CIFAR-10 using ResNet-18.}
 \setlength{\tabcolsep}{2pt}
 \resizebox{.75\textwidth}{!}
 {
  \begin{tabular}{c|ccccccc}
   \toprule[1.2pt]
         \# of transformations         &2 & 4&6&8&10&12&14\\ \hline
         Accuracy (\%)&95.99\scriptsize{$\pm$0.02}&96.18\scriptsize{$\pm$0.04}&96.38\scriptsize{$\pm$0.05}&96.54\scriptsize{$\pm$0.10}&96.61\scriptsize{$\pm$0.11}&96.67\scriptsize{$\pm$0.09} &\textbf{96.75}\scriptsize{$\pm$0.06} \\
   \bottomrule[1.2pt]
  \end{tabular}
  }
 \label{tab:augmentation-M}
\end{table*}
\noindent \textbf{Effectiveness of mitigating overfitting.} \quad
Although modern deep models can fit the distribution of training datasets through training, this capability often faces overfitting risks, which degrades models' generalization performance, particularly in scenarios with limited training data.
Data augmentation is primarily designed to address this issue by increasing the diversity of training data.
To evaluate the effectiveness of AdaAugment in mitigating overfitting risks, we simulate a high-overfitting-risk scenario by training a large ResNet-50 model on the relatively simple CIFAR-10 dataset. To further amplify the overfitting risk, we randomly reduce the sizes of the training set by 50\% while keeping the test set unchanged.

As can be seen in Figure~\ref{fig-overfitting-risk}, without AdaAugment, models achieve very high training accuracy, nearing 100\%, but the test accuracy rises slowly and exhibits significant fluctuations, indicating poor generalization and significant overfitting.
With AdaAugment, models achieve slower training accuracy growth and consistently lower training accuracy. Despite this, the test accuracy is significantly higher and more stable, demonstrating AdaAugment’s ability to mitigate overfitting and enhance generalization performance.
 
\begin{table}[]
	\centering
        \caption{Impact of adaptive magnitude $m$. Comparative analysis of AdaAugment compared to different settings of $m$ with CIFAR-100 using ResNet-18/50.}\label{tab:ablation}
	\renewcommand\arraystretch{1.1}
	\resizebox{.95\columnwidth}{!}{
		\begin{tabular}{c|ccccc}
			\toprule[1.5pt]
               Model & $m=0.5$ & random $m$ &linear $m$&sine $m$& AdaAugment \\ \hline
               R-18 & 78.62\scriptsize{$\pm$0.32} & 77.08\scriptsize{$\pm$0.30} &78.38\scriptsize{$\pm$0.27} &78.58\scriptsize{$\pm$0.32}&\textbf{79.84}\scriptsize{$\pm$0.27} \\
               R-50 & 80.23\scriptsize{$\pm$0.29} & 80.61\scriptsize{$\pm$0.19} &80.28\scriptsize{$\pm$0.31} &80.65\scriptsize{$\pm$0.15}&\textbf{81.46}\scriptsize{$\pm$0.12} \\
			\bottomrule[1.5pt]
		\end{tabular}}
\end{table}
\begin{table}[]
    \centering
        \caption{Effect of different RL modules with ResNet-18 on CIFAR-10/100.}\label{tab:rl-module-sac-ddpg}
	\renewcommand\arraystretch{1.1}
	\resizebox{.85\columnwidth}{!}{
		\begin{tabular}{c|ccc}
			\toprule[1.5pt]
               Dataset & DDPG~\cite{ddpg} & SAC~\cite{sac} & Ours \\ \hline
               CIFAR-10 &96.73\scriptsize{$\pm$0.06}&\textbf{96.82}\scriptsize{$\pm$0.04}&96.75\scriptsize{$\pm$0.06} \\
               CIFAR-100 &79.27\scriptsize{$\pm$0.19}&79.80\scriptsize{$\pm$0.29}&\textbf{79.84}\scriptsize{$\pm$0.27} \\
			\bottomrule[1.5pt]
		\end{tabular}}
\end{table}
\begin{table}[t]
	\centering
        \caption{Effect of $\gamma$ with CIFAR-100 using ResNet-18/50.}\label{tab:gamma}
	\renewcommand\arraystretch{1.1}
	\resizebox{.95\columnwidth}{!}{
		\begin{tabular}{c|ccccc}
			\toprule[1.5pt]
               $\gamma$ & 0.2&0.4&0.6&0.8&0.99 \\ \hline
               R-18 &78.94\scriptsize{$\pm$0.33}&78.95\scriptsize{$\pm$0.20}&78.99\scriptsize{$\pm$0.29}&79.24\scriptsize{$\pm$0.25}&\textbf{79.84}\scriptsize{$\pm$0.27} \\
               R-50 &80.61\scriptsize{$\pm$0.15}&80.44\scriptsize{$\pm$0.20}&80.68\scriptsize{$\pm$0.17}&80.90\scriptsize{$\pm$0.19}&\textbf{81.46}\scriptsize{$\pm$0.12} \\
            
			\bottomrule[1.5pt]
		\end{tabular}}
\end{table}
\begin{table}[h!]
	\centering
        \caption{Effect of $\lambda$ with CIFAR-100 using ResNet-18/50.}\label{tab:lambda}
	\renewcommand\arraystretch{1.1}
	\resizebox{.85\columnwidth}{!}{
		\begin{tabular}{c|ccc}
			\toprule[1.5pt]
               $\lambda$ & 0& 1 & Ours \\ \hline
            ResNet-18 &78.53\scriptsize{$\pm$0.36} &78.70\scriptsize{$\pm$0.30} &\textbf{79.84}\scriptsize{$\pm$0.27} \\
            ResNet-50 &80.21\scriptsize{$\pm$0.19} &80.74\scriptsize{$\pm$0.29} & \textbf{81.46}\scriptsize{$\pm$0.12}  \\
			\bottomrule[1.5pt]
		\end{tabular}}
\end{table}
\subsection{Ablation Study}
\noindent\textbf{Effect of the augmentation space.}\quad
Although the augmentation space used in AdaAugment is not larger than other methods~\cite{autoaugment,trivialaugment,randaugment,madaug,fast_autoaugment}, we further conduct analytical experiments to evaluate the impact of base augmentations.
Specifically, we analyze the performance by changing the number of augmentation transformations in our augmentation space.
We present the results in Table~\ref{tab:augmentation-M}.
Notably, the number of augmentation types used in Table~\ref{tab:augmentation-M} is consistently lower than other baseline approaches.
It can be observed that while the accuracy decreases as fewer augmentations are utilized, it drops very slowly.
When $N=2$, only two types of augmentations will be employed, which limits the diversity of augmented data.
Even under this case, the accuracy drops less than 0.8\%.
This highlights the effectiveness of AdaAugment in adaptively adjusting the augmentation magnitudes, effectively enhancing the model generalization.

\noindent\textbf{Effect of the RL module.}\quad
By utilizing the RL module to adaptively adjust the augmentation magnitude of DA operations, AdaAugment eliminates the need for manual parameter design and adjustment. 
In this section, we conduct experiments to assess the effect of the RL module.
By using the same augmentation space, we employ different experimental settings on $m$: fixed and random values of $m$, as well as increasing $m$ using linear and sine functions.
The results are summarized in Table~\ref{tab:ablation}.
It can be seen that the RL module brings substantial improvements in model performance.
By estimating the real-time model training status through the RL module, AdaAugment can optimize the augmentation magnitudes during training.
Thus, we demonstrate that the superior efficacy is brought by the adaptive adjustment of augmentation magnitude rather than the base augmentation space.

Additionally, in AdaAugment, we utilize lightweight A2C architectures. Here, we also employ two other more recent RL architectures, including Soft Actor-Critic (SAC)~\cite{sac} and Deep Deterministic Policy Gradient (DDPG)~\cite{ddpg}. As can be seen in Table~\ref{tab:rl-module-sac-ddpg}, the performance differences among different RL modules are relatively marginal.

\noindent\textbf{Effect of the discount factor $\gamma$.}\quad
Following previous works~\cite{gamma-1,gamma-2,gamma-3}, we set $\gamma$ as a constant 0.99. Here, we investigate the effect of different $\gamma$ values.
The results are presented in Table~\ref{tab:gamma}. It can be seen that AdaAugment performs best under our setting.

\noindent\textbf{Effect of parameter $\lambda$ in Eq.~\eqref{eq:reward}.}\quad
We conduct analytical experiments to evaluate the effect of parameter $\lambda$ in Eq.~\eqref{eq:reward}. Specifically, instead of dynamically adjusting $\lambda$, $\lambda$ is fixed as 0 and 1. The results are shown in Table~\ref{tab:lambda}, validating the stability of parameter $\lambda$.

\section{Discussion and Future Work}
In this work, we propose AdaAugment, a tunning-free adaptive data augmentation method. Instead of relying on any hand-crafted criteria, AdaAugment incorporates an RL module to adaptively adjust the augmentation magnitudes for individual training samples. This adaptive adjustment aligns with the model training progress.
Experiment results demonstrate that AdaAugment achieves enhanced model performance across various benchmark datasets without introducing noticeable training overhead.
Besides strengths, the limitations of our method should also be discussed.

While the proposed method estimates the risks of underfitting and overfitting by deriving losses from fully augmented and nonaugmented data, this additional loss calculation requires two extra forward passes compared to vanilla training.
To enhance the overall efficiency, our policy network is designed to be consistent across various target models.
Thus, the increase in GPU hours is relatively modest and not directly correlated with the complexity of the task model. 
Future work should consider leveraging other surrogate indicators to estimate underfitting and overfitting risks.

In our augmentation space, we leverage image transformations whose applied strengths can be adjusted with a real-valued parameter, i.e., $m \in [0,1]$.
However, most existing DA methods, such as Cutout~\cite{cutout}, AdvMask~\cite{advmask}, and AutoAugment~\cite{autoaugment}, do not support explicit adjustments of augmentation magnitudes. 
Therefore, future work should consider exploring an adjustable version of existing advanced DA methods, facilitating explicit adjustment of augmentation strength with continuous real values. 
Consequently, our augmentation space may be further expanded.
This expansion holds promise for further performance enhancements.

\section{Conclusion}
In this work, we propose AdaAugment, an innovative and tuning-free method for adaptive data augmentation.
By adaptively adjusting the magnitudes of DA operations based on the real-time feedback obtained during the training of the target model, AdaAugment optimizes the data variability.
This dynamic adjustment ultimately mitigates the risks associated with both underfitting and overfitting, thereby optimizing model training and enhancing model generalization capabilities.
Through extensive empirical evaluations, we demonstrate that our adaptive augmentation mechanism helps significantly improve performance across various benchmark datasets with competitive efficiency.

\section*{Acknowledgments}
This work is supported by the STI 2030-Major Projects of China under Grant 2021ZD0201300, the Fundamental Research Funds for the Central Universities under Grant 2024300394, the National Natural Science Foundation of China under Grant 62276127.
\bibliographystyle{IEEEtran}
\bibliography{ref}  
\end{document}